\pdfoutput=1
\documentclass{article}

\usepackage[preprint]{neurips_2022}

\usepackage{natbib}
\setcitestyle{numbers,square}
\usepackage[utf8]{inputenc} 
\usepackage[T1]{fontenc}    
\usepackage{hyperref}       
\usepackage{url}            
\usepackage{booktabs}       
\usepackage{amsfonts}       
\usepackage{nicefrac}       
\usepackage{microtype}      
\usepackage{xcolor}         
\usepackage{amsmath}
\usepackage{amssymb}
\usepackage{mathtools}
\usepackage{amsthm}
\usepackage{multirow}
\usepackage{wrapfig,lipsum,booktabs}
\usepackage{caption}
\usepackage{ulem}
\newcommand{\tabincell}[2]{\begin{tabular}{@{}#1@{}}#2\end{tabular}}

\title{Bridging the Gap Between Training and Inference of Bayesian Controllable Language Models}

\author{
  Han Liu\\
  Tsinghua University\\
  \texttt{han-liu18@mails.tsinghua.edu} \\
  \And
  Bingning Wang \\
  Tencent Inc. \\
  \texttt{bryantwwang@tencent} \\
  \And
  Ting Yao \\
  Tencent Inc. \\
  \texttt{tessieyao@tencent} \\
  \AND
  Haijin Liang \\
  Tencent Inc. \\
  \texttt{hodgeliang@tencent} \\
  \And
  Jianjin Xu \\
  Tsinghua University \\
  \texttt{xujj15@gmail} \\
  \And
  Xiaolin Hu \\
  Tsinghua University \\
  \texttt{xlhu@tsinghua.edu} \\
}

\begin{document}

\maketitle

\begin{abstract}
Large-scale pre-trained language models have achieved great success on natural language generation tasks. However, it is difficult to control the pre-trained language models to generate sentences with the desired attribute such as topic and sentiment, etc. Recently, Bayesian Controllable Language Models (BCLMs) have been shown to be efficient in controllable language generation.
Rather than fine-tuning the parameters of pre-trained language models, BCLMs use external discriminators to guide the generation of pre-trained language models.
However, the mismatch between training and inference of BCLMs limits the performance of the models. 
To address the problem, in this work we propose a ``Gemini Discriminator'' for controllable language generation which alleviates the mismatch problem with a small computational cost.
We tested our method on two controllable language generation tasks: sentiment control and topic control. On both tasks, our method reached achieved new state-of-the-art results in automatic and human evaluations.
\end{abstract}

\section{Introduction}
\label{sec:introduction}
Recent advances in large-scale pre-trained language model (PLM) \cite{radford2017learning,radford2018improving,radford2019language,GPT3} have made great progress on natural language processing (NLP) tasks. One of the most attractive pretraining paradigms is auto-regressive language model such as GPT2~\cite{radford2019language} and XLNet~\cite{yang2019xlnet}. With billions or even trillions of parameters, and abundant unlabeled training data, PLMs can generate diverse and realistic sentences.  
Formally, autoregressive PLM models the probability distribution of text $X=\{x_1,x_2,...,x_T\}$ with the chain rule:
\begin{equation}
\begin{aligned}
p(X)=\prod_{i=1}^{T}p(x_{i}|x_1,x_2,...,x_{i-1}).
\label{equ:lm_p}
\end{aligned}
\end{equation}
However, those models are usually trained on general purpose corpus and the sentences generated by those PLMs are usually inconsistent with task requirements. 
Therefore, to improve the applicability of PLMs, making the PLMs more controllable has been an important task in natural language generation.
Controllable language generation attempts to model $p(X|a)$ where $a$ is a desired attribute (e.g. topic, length and sentiment):
\begin{equation}
\begin{aligned}
p(X|a)=\prod_{i=1}^{T}p(x_{i}|X_{1:i-1},a).
\label{equ:clm_p}
\end{aligned}
\end{equation}
To simplify the expression, we use $X_{1:i}$ to denote the sequence $\{x_1,x_2,...,x_i\}$.
Earlier works directly model $p(X|a)$ by maximizing the likelihood of the task-specific corpus. These methods are called \textit{Class Conditional Language Models} (CCLMs)~\cite{SeqGAN,Hu2017Toward,keskarCTRL2019}.
Since the training data for specific attributes is limited, CCLMs usually suffer from corpus overfitting.

\textit{Bayesian Controllable Language Models} (BCLMs) \cite{dathathri2019plug,krause-etal-2021-gedi-generative,yang2021fudge} are proposed to solve the corpus overfitting problem. They attempt to find a generation path of the pre-trained language model that matches the target attribute. 
BCLMs use a discriminator to model the class probability $p(a|X_{1:i})$ and then sample the desired output from $p(x_{i}|X_{1:i-1},a)$ according to the Bayes Rule:
\begin{equation}
\begin{aligned}
p(x_{i}|X_{1:i-1},a)\propto p(a|X_{1:i})p(x_{i}|X_{1:i-1}).
\label{equ:bayes_p}
\end{aligned}
\end{equation}
With the Bayes Rule, BCLMs have converted the controllable language generation task to the combination of a unconditional language generation task and a classification task. Since the $p(x_{i}|X_{1:i-1})$ can be given by an off-the-shelf PLM (in this paper we always use the GPT2 model), BCLMs only need to model $p(a|X_{1:i})$. 

BCLMs usually train the discriminator with the sentences from the task-specific corpus to model $p(a|X_{1:i})$. However, in inference, the discriminator receives the text generated by GPT2 as input which is different from the training corpus. The mismatch between training and inference limits the performance of BCLMs.

In this work, we propose to use GPT2 to extract the features of the sentences from the task-specific corpus and use the features as the input to the discriminator. Since GPT2 has been pre-trained on general purpose corpus, it is reasonable to assume that it has ``seen'' similar sentences from the task-specific corpus. It is roughly equivalent to saying that the discriminator has been pre-trained on the large-scale general-purpose corpus. This mitigates the mismatch problem in BCLMs. 

However, doing this job is challenging. In Equation~\eqref{equ:bayes_p}, $x_{i}$ is to be generated at step $i$ in inference. To get the probability distribution of $x_{i}$, we need to calculate $p(a|X_{1:i-1}, w)$ for all token $w$ in the vocabulary. It is a huge cost for extracting the features of sentence $\{X_{1:i-1}, w\}$ for all token $w$ in the vocabulary.

To address the efficiency problem, we introduce ``Gemini Discriminator'' (Gemini) for controllable language generation.
The Gemini uses the features extracted by GPT2 as input in training, and has small computational cost in inference.
To improve the performance of Gemini, we adopt knowledge distillation~\cite{distilling,kim-rush-2016-sequence} in training.
Moreover, inspired by the nucleus sampling \cite{topp} and the decoding strategy in GeDi \cite{krause-etal-2021-gedi-generative}, we design an attribute-driven nucleus sampling method in generation, which considers both the fluency and the attribute relevance in generation.

We experimented on two controllable language generation tasks: sentiment control and topic control. On both tasks, Gemini reached new state-of-the-art results in both automatic and human evaluations. 

\section{Related Work}
Controllable language models can be classified into two categories: CCLM and BCLM.

\textbf{CCLM:} CCLM is the most straightforward approach in controllable language generation which directly models the probability $p(X|a)$. There are various methods to implement CCLM, such as training conditional generative models by concatenating the desired attribute to inputs \cite{ficler-goldberg-2017-controlling,kikuchi-etal-2016-controlling}, training generative adversarial networks \cite{SeqGAN}, and training variational auto-encoders \cite{Hu2017Toward}. In recent years, with the development of large-scale pre-trained language models \cite{radford2017learning,radford2018improving,radford2019language,GPT3} and prompt methods \cite{jiang2020can,Li2021PrefixTuningOC,Shin2020Auto}, great progress has been made in CCLMs. CTRL \cite{keskarCTRL2019} builds a large controllable language model trained on a large-scale corpus with 55 different control codes. Following CTRL, COCON \cite{chan2020cocon} proposes three self-supervised learning loss to enhance controllability. Moreover, ACB \cite{yu2021attribute} tries to disentangle the irrelevant attributes in corpus and introduces prefix-tuning in CCLM to avoid tuning a large number of parameters. 

Since the training data for specific attributes is limited, CCLMs usually lead to corpus overfitting, meaning that the generated sentences of CCLMs usually highly resemble sentences in the training corpus. For example, sentences sampled from CCLMs with sentiment control trained on IMDB \cite{maas-etal-2011-imdb}, a dataset of movie reviews for sentiment classification, usually mention words in movie reviews (e.g. movie, film, and character). However, the attribute we wish CCLMs to learn is the sentiment in IMDB rather than the context relevance of movie reviews.

\textbf{BCLM:}
BCLMs attempt to find the generation path that matches the target attribute by using an external discriminator. 
Instead of modeling $p(X|a)$ directly, BCLMs sample $X$ according to the Bayes Rule: $p(X|a)\propto p(a|X)p(X)$. 
PPLM \cite{dathathri2019plug} proposes an iterative method that updates the hidden states of a language model by backpropagating gradients of an attribute model to maximize the likelihood of the desired attribute. Since PPLM requires iterations of forward and backward processes in inference, it is computationally intensive.
GeDi \cite{krause-etal-2021-gedi-generative} uses smaller language models to model $p(a|X)$.
FUDGE \cite{yang2021fudge} trains a discriminator to distinguish whether the desired attribute will be true in the future and use the discriminator to model $p(a|X)$.
However, as discussed in Section~\ref{sec:introduction}, the mismatch between training and inference of the discriminator limits the performance of BCLMs.

\begin{figure*}[!t]
  \centering
  \includegraphics[width=\textwidth]{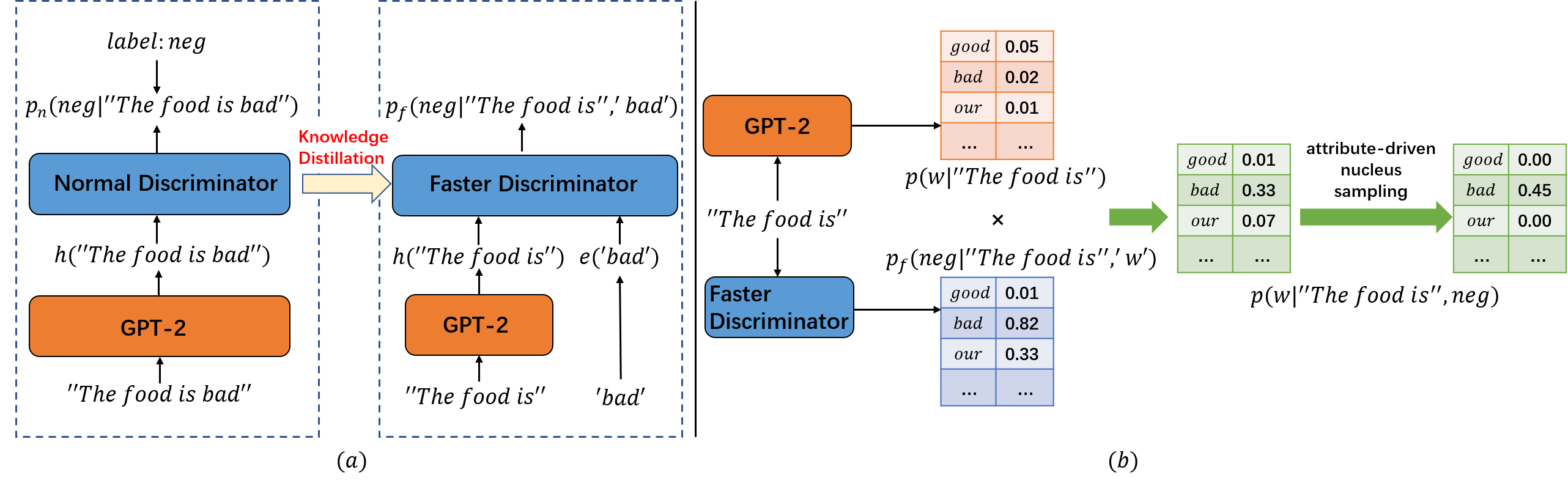}
  \caption{\small{Illustration of the Gemini method. (a) The sketch of Gemini training. Knowledge distillation is used to train the faster discriminator. (b) The sketch of Gemini inference. We use the faster discriminator together with the GPT2 in inference and adopt the attribute-driven nucleus sampling as the decoding strategy.}}
\label{fig:wem_intro}
\end{figure*}

\section{Methods}
\label{sec:method}
\subsection{Gemini Training}
Figure~\ref{fig:wem_intro} (a) demonstrates the sketch of Gemini training.
Similar to the twins of Gemini in ancient Greek mythology, the ``Gemini Discriminator'' has two modules: a normal discriminator and a faster discriminator.
Same as previous BCLMs,
Gemini models $p(x_{i}|X_{1:i-1})$ in Equation~\eqref{equ:bayes_p} by a pre-trained GPT2 model $G$. To avoid corpus overfitting, all of the parameters in $G$ are fixed in training. 
Calculating $p(a|X_{1:i})$ is modeled as a classification task. 
To simplify the problem, we treat the classification task as a binary classification task. 
We label the sentences in the training corpus by $y\in \{0,1\}$ ($y=1$ when the sentence contains attribute $a$, and $y=0$ otherwise).
For convenience, in the following, we denote the last hidden state of $G$ as $h_1,...,h_T$ with $X_{1:T}$ as the input. 

Given $h_{i}$, the normal discriminator adopts a neural network $M_{\texttt{n}}$ to classify sentence $\{X_{1:i},w\}$. Specifically we use the output of $M_{\texttt{n}}$ with a sigmoid function $\sigma$ to model $p(a|X_{1:i})$:
\begin{equation}
\begin{aligned}
&p_{\texttt{n}}(a|X_{1:i})=\sigma(M_{\texttt{n}}(h_{i})).
\label{equ:normal discriminator}
\end{aligned}
\end{equation} 
we train the normal discriminator by minimizing the cross-entropy loss. The average probability over time is used in the cross-entropy loss since in experiments we found that the average probability could make the training more smooth. The loss function for the normal discriminator is
\begin{equation}
\begin{aligned}
L_{\texttt{xe}}&=-y\ln p_{\texttt{avg}} -(1-y)\ln(1-p_{\texttt{avg}}), \\
p_{\texttt{avg}}&=-\frac{1}{T}\sum_{i=1}^{T}p_{\texttt{n}}(a|X_{1:i}).
\label{equ:xe loss}
\end{aligned}
\end{equation}

As discussed in Section~\ref{sec:introduction}, using $h_{i}$ to model $p(a|X_{1:i})$ will lead to a huge time cost in inference since we need to extract the features of $\{X_{1:i-1},w\}$ for all token $w$ in the vocabulary. 
We propose a faster discriminator that will be used in inference, and use the normal discriminator to teach the faster discriminator.

The faster discriminator predicts $p(a|X_{1:i-1},w)$ according to $h_{i-1}$ and $e_{w}$, where $e_{w}$ is the pre-trained word embedding of $w$ and is fixed in training. With the faster discriminator, we only need to forward GPT2 once for extracting the features for the input $X_{1:i-1}$, which significantly reduces the computational cost.
In the faster discriminator, given $h_{i-1}$ and a token $w$ in the vocabulary, a neural network $M_{\texttt{f}}$ is used to classify $\{X_{1:i-1},w\}$. Specifically, we use the output of $M_{\texttt{f}}$ with a sigmoid function $\sigma$ to model $p(a|X_{1:i-1},w)$:
\begin{equation}
\begin{aligned}
&p_{\texttt{f}}(a|X_{1:i-1},w)=\sigma(M_{\texttt{f}}(h_{i-1},e_{w})).
\label{equ:predict module}
\end{aligned}
\end{equation}

If we replace the token $w$ in $p_{\texttt{f}}(a|X_{1:i-1},w)$ with $x_{i}$ in the training sample, $p_{\texttt{f}}(a|X_{1:i-1},x_{i})$ will be identical to $p_{\texttt{n}}(a|X_{1:i})$. The connection between the normal discriminator and the faster discriminator permits us to use knowledge distillation to train the faster discriminator. Concretely, the loss function for training the faster discriminator is
\begin{equation}
\begin{aligned}
L_{\texttt{kd}}=\frac{1}{T-1} \sum_{i=2}^{T}||M_{\texttt{n}}(h_{i})-M_{\texttt{f}}(h_{i-1},e_{x_{i}})||^2 .
\label{equ:kd loss}
\end{aligned}
\end{equation}
We use the Euclidean distance between $M_{\texttt{n}}$ and $M_{\texttt{f}}$ rather than the distance between $p_{\texttt{n}}$ and $p_{\texttt{f}}$ to avoid the vanishing gradient problem of the sigmoid function.

The normal discriminator and the faster discriminator are trained at the same time. The final loss is
\begin{equation}
\begin{aligned}
L_{\texttt{final}}=L_{\texttt{xe}}+L_{\texttt{kd}}.
\label{equ:final loss}
\end{aligned}
\end{equation}

\subsection{Gemini Inference}
\label{sec:decoding strategy}
Figure~\ref{fig:wem_intro} (b) demonstrates the sketch of Gemini inference. We use the class probability predicted by the faster discriminator to guide the generation of $G$ in inference. 
At inference step $i$, the faster discriminator calculates $p_{\texttt{f}}(a|X_{1:i-1},w)$ for each token $w$ in the vocabulary.
According to the Bayes Rule, we can sample sequences conditioned on $a$ with $p(w|X_{1:i-1},a)\propto p_{\texttt{f}}(a|X_{1:i-1},w)p(w|X_{1:i-1})$. 

Following previous BCLM works~\cite{dathathri2019plug,krause-etal-2021-gedi-generative}, We applies weighted decoding in our model by introducing a non-negative hyper-parameter $\lambda$. The conditioned probability with weighted decoding is
\begin{equation}
\begin{aligned}
p(w|X_{1:i-1},a)\propto p_{\texttt{f}}(a|X_{1:i-1},w)^{\lambda}p(w|X_{1:i-1}).
\label{equ:weighted decoding}
\end{aligned}
\end{equation}
The larger $\lambda$ is, the more attention we pay to controllability. When $\lambda$ is $0$, the conditioned probability degrades into the unconditioned probability.

Moreover, we design an attribute-driven nucleus sampling in inference to improve the controllability and the linguistic quality. 
The attribute-driven nucleus sampling contains two filters. The first filter is a standard nucleus sampling filter~\cite{topp} acting on the unconditional probability distribution $p(w|X_{1:i-1})$ to maintain the linguistic quality of generation. We define $V_k$ as the set of $k$ tokens $w$ with the highest $p(w|X_{1:i-1})$. With a filter probability $\rho_1$, $\hat{k}(\rho_1)$ is defined as
\begin{equation}
\begin{aligned}
\hat{k}(\rho_1) = \mathop{\texttt{argmin}}\limits_{k} (\sum_{w\in V_k}p(w|X_{1:i-1})\ge \rho_1).
\label{equ:filter1}
\end{aligned}
\end{equation}
With the first filter, we get the $\hat{k}(\rho_1)$ tokens with the highest $p(w|X_{1:i-1})$ which form a set $V_{\hat{k}(\rho_1)}$. 
Inspired by the decoding strategy of GeDi~\cite{krause-etal-2021-gedi-generative},
the second filter, named ``attribute-driven filter'',  acts on the conditional probability distribution $p(w|X_{1:i-1},a)$, aiming at making the generated sentences well conditioned on the attribute $a$. We define $U_m$ as the set of $m$ tokens $w$ in $V_{\hat{k}(\rho_1)}$ with the highest $p_{\texttt{f}}(a|X_{1:i-1},w)$. With a filter probability $\rho_2$, $\hat{m}(\rho_2)$ is defined as
\begin{equation}
\begin{aligned}
\hat{m}(\rho_2)=\mathop{\texttt{argmin}}\limits_{m}(\sum_{w\in U_m}p(w|X_{1:i-1},a)\ge \rho_2).
\label{equ:filter2}
\end{aligned}
\end{equation}
Finally we sample from the set $U_{\hat{m}(\rho_2)}$. 
Experimentally, we set $\rho_1=0.9$, $\rho_2=0.3$ in all experiments.

 \section{Experiments}
We tested Gemini on two controllable language generation tasks: sentiment control and topic control. 

\subsection{Datasets}
\label{sec:dataset}
\textbf{IMDB:} We used the IMDB dataset \cite{maas-etal-2011-imdb} for sentiment control. It contains 50k samples of movie reviews labeled with sentiment tag (25k for positive sentiment and 25k for negative sentiment). 
We trained our model on both positive attribute and negative attribute. In evaluation, we used the same 15 prefixes (See Section~\ref{sec:prefixes}) of sentiment control from the prior work \cite{dathathri2019plug} and generated samples started with these prefixes. 

\textbf{AG News:} We used the AG News dataset \cite{agnews} for topic control. The AG News dataset contains 120k samples of news articles. The samples are classified into one of the 4 classes: worlds, sports, business, and science. We experimented on all of the 4 classes. 
In evaluation, we used the same 20 prefixes (See Section~\ref{sec:prefixes}) of topic control from the prior work \cite{dathathri2019plug} and generated samples from these prefixes. 

\subsection{Baselines}
\label{sec:baselines}
We choose the models below as the baseline models.
\textbf{GPT2-ranked} \cite{radford2019language} is the unconditional language model pre-trained on large-scale unlabeled corpus with an attribute classifier. The nucleus sampling~\cite{topp} was used in inference with a filter probability 0.7. We sampled a GPT2 3 times, then used the best sample ranked with the likelihood of a BERT classifier as the generated results. 
\textbf{GPT2-tune} is the GPT2 fine-tuned on the corpus with desired attributes. The prefix-tuning \cite{Li2021PrefixTuningOC} was used in the GPT2-tune model. The length of prompt in prefix-tuning was set to 3. Same as GPT2-ranked, nucleus sampling was applied in inference with a filter probability 0.7.
\textbf{PPLM} \cite{dathathri2019plug} is a BCLM method that iteratively updates the hidden states of GPT2 in inference. Following the settings in the original paper of PPLM, we set the iteration steps of PPLM to 5 and used a top-k sampling \cite{topk} with $k=10$ in inference.
\textbf{GeDi} \cite{krause-etal-2021-gedi-generative} is a recent BCLM model which models the probability of desired and undesired attributes by CCLMs together with the Bayes Rule.  
\textbf{FUDGE} \cite{yang2021fudge} is also a recent BCLM model that uses
an LSTM~\cite{hochreiter1997long} as the discriminator to predict whether the attribute will be achieved in future generation. Since the original paper on FUDGE didn't experiment on the IMDB and the AG News datasets, we used the FUDGE model implemented by ourselves instead.

\subsection{Implementation Details}
We used the GPT2-large implemented in Huggingface Transformers\footnote{https://huggingface.co/gpt2-large} \cite{wolf2020transformers} as the backbone GPT2 model.
The AdamW optimizer \cite{loshchilov2017decoupled} was adopted in all experiments. The learning rate of AdamW was set to $5\times 10^{-5}$.
For Gemini experiments, we used a batch size 64 and trained the models for 200 epochs. On average, on a Nvidia A100 Tensor Core GPU machine, each epoch took 7 minutes and 8 minutes for sentiment control and topic control respectively. For GPT2-tune experiments, we used a batch size 16 and trained the models for 50 epochs. 
In inference, the $\lambda$ in Equation \eqref{equ:weighted decoding} was set to 5.0 in both sentiment control and topic control experiments.

\subsection{Automatic Evaluation}
\label{sec:automatic evaluation}
In controllable language generation, we need to evaluate both the attribute relevance and the linguistic quality of the generated sentences. Moreover, the diversity and the resemblance between the generated sentences and the training corpus are also important evaluation metrics. In automatic evaluation, for each model, we generated 50 samples per prefix and report the average score over the samples.

\textbf{Attribute relevance (AR):}
For both tasks, we used classifiers based on BERT \cite{bert} to measure the consistency between the generated sentences and the desired attribute. 
The BERT was initialized with the BERT-large implemented in Huggingface Transformers\footnote{https://huggingface.co/bert-large-uncased} and fine-tuned on the task-specific corpus.
\textbf{Perplexity (PPL):} Since we used GPT2-large as the backbone of our method and baselines, for the sake of fairness, we measured the linguistic quality of generated sentences by the average perplexity of a GPT2-XL\footnote{https://huggingface.co/gpt2-xl}. 
\textbf{Excellent rate (ER):} Since there is a trade-off between controllability and linguistic quality, we propose the excellent rate, an evaluation metric that considers both of them. The excellent rate is defined as the rate of the sample with more than 90\% attribute relevance likelihood as well as less than 40 perplexity. We found the generated sentences that met the requirement were usually both fluent and relevant to the attribute in human evaluation. 
\textbf{Diversity:} We report the distinct 1-2-3-grams (a.k.a. \textbf{Dist-1}, \textbf{Dist-2}, \textbf{Dist-3}) \cite{li-etal-2016-diversity} for diversity evaluation.
\textbf{Corpus resemblance (CR):} We generated 30000 sentences from the GPT2-large and randomly chose 10000 samples from the task-specific corpus, then fine-tuned a BERT-large to classify the source of the sentences. The corpus resemblance is the average likelihood calculated by the BERT-large over the generated sentences.
This metric is used to evaluate the resemblance between the generated sentences and the training corpus. Lower corpus resemblance is better in the controllable language generation tasks. 

\subsection{Human Evaluation}
In human evaluation, for each model we generated 5 sentences per prefix.
We hired 15 annotators for sentiment classification (7 females and 8 males), and 20 annotators for topic classification (8 females and 12 males). The age of annotators varied between 22 and 30, and the degree of annotators varied from undergraduate to doctorate. 

\textbf{Evaluation of sentiment control:} the generated sentences are scored on sentiment relevance, linguistic quality and resemblance with movie reviews. All metrics are on 1-5 Likert scale. 
For sentiment classification, annotators were asked to evaluate the positivity of samples with 1 for very negative and 5 for very positive. 
We use the score of the positivity of samples generated by positive control and 6 minus the score of the positivity of samples generated by negative control as the attribute relevance score. We report the average attribute relevance (AR) score over all samples.
For linguistic quality (LQ) evaluation, annotators measured whether the samples were coherent, fluent, and understandable, then scored the samples with 1 for totally not fluent and 5 for very fluent. 
For corpus resemblance (CR), annotators were asked to measure whether the samples were like movie reviews, then scored the samples with 1 for low similarity and 5 for high similarity.

\textbf{Evaluation of topic control:} the generated sentences are scored on attribute relevance (AR) and linguistic quality (LQ). For attribute relevance, annotators were asked to evaluate the degree of relevance between the generated sentences and the desired topic. Evaluation of linguistic quality in topic control was same as that in the sentiment control.
\begin{table*}[!t]
\small
\caption{\small{Examples generated by Gemini. Words underscored are the prefixes. Words that signify the controlled effect are highlighted in red.} }
\begin{center}
\resizebox{\textwidth}{28mm}{
\begin{tabular}{l|l}
\toprule
Attribute & Generated Sentence\\
\midrule
Positive & \tabincell{l}{\uline{The painting} is a \textcolor{red}{masterpiece}, and a \textcolor{red}{true} work of art, and a \textcolor{red}{great} work of visual art in the sense that \\it's a \textcolor{red}{beautiful} representation of the complexities of \textcolor{red}{love} and its dynamic relationship with our own lives.  } \\
\midrule
Negative & \tabincell{l}{\uline{The painting} seemed \textcolor{red}{pretty stupid}, what \textcolor{red}{a waste of money} and effort you spent to make this kind \\of \textcolor{red}{stupid trash}. I'm sorry, but I just can't believe how \textcolor{red}{dumb} this whole thing was.} \\
\midrule
World & \tabincell{l}{\uline{This essay discusses} \textcolor{red}{President Barack Obama's} State of the \textcolor{red}{Union speech} on \textcolor{red}{immigration} and \\\textcolor{red}{deportations} in the House and Senate during his second term in office.}\\
\midrule
Sports & \tabincell{l}{\uline{This essay discusses} \textcolor{red}{Michael Phelps'} \textcolor{red}{hockey} career and the professional \textcolor{red}{hockey player's} peak.\\His \textcolor{red}{NHL} career was won in a \textcolor{red}{Game 7} win against the \textcolor{red}{Chicago Blackhawks} in the 2006 \textcolor{red}{Olympic Games}.} \\
\midrule
Business & \tabincell{l}{\uline{This essay discusses} \textcolor{red}{financial savings} and \textcolor{red}{financial markets}, the US stock \textcolor{red}{market}, \textcolor{red}{Treasury debt} \\securities, interest rates and the Federal Reserve's \textcolor{red}{monetary policy}. } \\
\midrule
Science & \tabincell{l}{\uline{This essay discusses} \textcolor{red}{open source software} and \textcolor{red}{Windows applications} which includes \textcolor{red}{Microsoft} \\\textcolor{red}{Windows application development tools} and \textcolor{red}{Windows desktop applications}.}\\
\bottomrule
\end{tabular}}
\end{center}
\label{tab:examples}
\end{table*}
\begin{table*}[!t]
\small
\caption{\small{Results on sentiment control task.} }
\begin{center}
\resizebox{\textwidth}{30mm}{
\begin{tabular}{l|ccc|ccccccc}
\toprule
\multirow{2}{*}{Model} &\multicolumn{3}{c|}{Human evaluation} &\multicolumn{7}{c}{Automatic evaluation} \\
& AR$\uparrow$ & LQ $\uparrow$ & CR$\downarrow$ &  AR(\%)$\uparrow$ & PPL$\downarrow$ & ER(\%)$\uparrow$ & Dist-1$\uparrow$ & Dist-2$\uparrow$ & Dist-3$\uparrow$ & CR(\%)$\downarrow$\\
\midrule
\multicolumn{11}{l}{\textit{Baseline CCLMs}} \\
\midrule
GPT2-ranked & 3.36 & 3.92 & 1.84 & 75.7 & 11.57 & 24.4 & 0.38 & 0.82 & \textbf{0.93} & 4.56 \\
GPT2-tune & 3.53 & \textbf{3.99} & 3.63 & 73.3 & 13.61 & 50.6  &0.33 &0.75 &0.87 & 56.79 \\
\midrule
\multicolumn{11}{l}{\textit{Baseline BCLMs}} \\
\midrule
PPLM & 3.66 & 2.77 & \textbf{1.68} & 63.9 & 16.95 & 23.4  &0.32 &0.67 &0.82 & 1.34\\
GeDi & 3.67 & 2.89 & 1.88 & 95.2 & 75.82 & 13.2 & \textbf{0.55} & \textbf{0.88} &0.90 & 5.48\\
FUDGE & - & - & - & 70.9 & \textbf{11.06} & 18.1 & 0.39 & 0.79 & 0.89 & \textbf{1.19}\\
\midrule
\multicolumn{11}{l}{\textit{Models for Ablation Study}} \\
\midrule
BCLM-GRU &- &-&-& 96.5 & 38.17 & 34.8 & 0.41& 0.78& 0.89 & 5.99 \\
Gemini-(w/o)KD &- &-&-& 90.9 & 44.62 & 31.5 & 0.45& 0.81 & 0.88 & 4.32 \\
Gemini-(w/o)AD &- &-&-& 75.4 & 18.02 & 22.0 & 0.43& 0.83& 0.91 & 1.82\\
\midrule
\midrule
Gemini &\textbf{4.16} & 3.50 & 1.96 & \textbf{98.7} & 18.99 & \textbf{75.3} & 0.31& 0.69& 0.85 & 10.36 \\
\bottomrule
\end{tabular}}
\end{center}
\label{tab:sentiment}
\end{table*}
\begin{table*}[!t]
\small
\caption{\small{Results on topic control task.} }
\begin{center}
\resizebox{\textwidth}{26mm}{
\begin{tabular}{l|cc|ccccccc}
\toprule
\multirow{2}{*}{Model} &\multicolumn{2}{c|}{Human evaluation} &\multicolumn{7}{c}{Automatic evaluation} \\
& AR$\uparrow$ & LQ $\uparrow$ & AR(\%)$\uparrow$ & PPL$\downarrow$ & ER(\%)$\uparrow$ & Dist-1$\uparrow$ & Dist-2$\uparrow$ & Dist-3$\uparrow$ & CR(\%)$\downarrow$\\
\midrule
\multicolumn{10}{l}{\textit{Baseline CCLMs}} \\
\midrule
GPT2-tune & 4.13 & \textbf{4.05} & 89.8 & 22.56 & 72.8 & 0.38 & 0.78 & 0.87 & 69.17 \\
\midrule
\multicolumn{10}{l}{\textit{Baseline BCLMs}} \\
\midrule
GeDi & 4.26 & 3.40 & 98.1 & 76.77 & 28.5 & \textbf{0.51} & \textbf{0.86} & 0.89 & 65.25\\
FUDGE & - & - & 85.1 & 14.52 & 77.8 & 0.38 &0.79 & 0.88 & \textbf{7.53} \\
\midrule
\multicolumn{10}{l}{\textit{Models for Ablation Study}} \\
\midrule
BCLM-GRU & - & - & 96.5 & 26.90 & 81.6 & 0.34 & 0.73 & 0.86 & 32.04 \\
Gemini-(w/o)KD & - & - & 98.2 & 24.51 & 85.0 & 0.42 &0.79 &0.88 & 17.31 \\
Gemini-(w/o)AD& - & - & 99.2 & \textbf{12.43} & 95.2 & 0.38 & 0.78 & \textbf{0.90} & 12.74 \\
\midrule
\midrule
Gemini & \textbf{4.62} & 3.79 & \textbf{99.6} & 16.18 & \textbf{97.8} & 0.27 & 0.64 &0.85 &27.85 \\
\bottomrule 
\end{tabular}}
\end{center}
\label{tab:topic}
\end{table*}

\section{Results} 
\subsection{Comparison with the Baseline Models}
\label{sec:comparison with baseline}
We demonstrate some examples generated by Gemini in Table~\ref{tab:examples}.
As shown in Table~\ref{tab:examples}, most of the sentences generated by Gemini are fluent and relevant to the desired attribute.
In our experiments, we found that the sentences generated by
PPLM contained a lot of repeated words, and the sentences generated by GeDi were usually incoherent. These problems decreased the linguistic quality of PPLM and GeDi.
Although the linguistic quality of GPT2-tune was very high, the generated sentences of the GPT2-tune highly resembled the movie review corpus. (Table~\ref{tab:sentiment samples} and~\ref{tab:topic samples})

Table~\ref{tab:sentiment} and Table~\ref{tab:topic} show the results of sentiment control and topic control respectively.

In the sentiment control task, the attribute relevance of Gemini in automatic evaluation reached 98.7\%, and in human evaluation the score of attribute relevance was 4.16.  
The attribute relevance scores of Gemini outperformed all the baseline models.
The linguistic quality score of Gemini was 3.50 in human evaluation, which outperformed all the BCLM baselines, though not as good as the CCLMs. 
Moreover, Gemini reached 75.3\% on the excellent rate, which exceeded the best of the baselines by 25 percentage points.

In the topic control task, Gemini also outperformed all the baseline models on attribute relevance (99.6\% in automatic evaluation and 4.62 in human evaluation), while maintaining good linguistic quality and low corpus resemblance. On the excellent rate, Gemini reached 97.8\%, which exceeded the best of the baselines by 20 percentage points.

However, Gemini performed poorly on diversity compared to the baseline models. The problem on diversity might be caused by our decoding strategy. See discussion in Section~\ref{sec:ablation study}. 

Moreover, we tested the inference speed of Gemini and the baseline models. Specifically, for each model, we generated 1000 samples with batch size 50 and calculated the average time per token. We report the mean generation time and the standard deviation in Table \ref{tab:time test}. All experiments were on a Nvidia A100 Tensor Core GPU machine. The inference speed of our Gemini method was about 13$\times$ faster than PPLM and about 2.5$\times$ faster than GeDi, though not as fast as the CCLMs (i.e. unconditional GPT2 and GPT2-tune).

Comparing the results in Table~\ref{tab:sentiment} with the results in Table~\ref{tab:topic}, we find that the attribute relevance in topic control task is generally better than that in sentiment control task. 
Compared with topic control task, sentiment control is more difficult since there are a lot of implicit expressions of sentiment and the sentiment of a sentence has a strong correlation with the context. 

\begin{minipage}[t]{\textwidth}
\begin{minipage}[t]{0.5\textwidth}
\makeatletter\def\@captype{table}
\small
\setcaptionwidth{0.9\textwidth}
\caption{\small{Time cost per token in inference. Mean$\pm$standard deviation.}}
\centering
\begin{tabular}{lc}
\toprule
Model   & Generation Time (s) \\
\midrule
GPT2       & 0.030$\pm$0.002      \\
GPT2-tune            & 0.030$\pm$0.008       \\
\midrule
PPLM    & 1.028$\pm$0.130      \\
GeDi   & 0.193$\pm$0.090       \\
FUDGE(our implem.)   & 0.030$\pm$0.009       \\
\midrule
Gemini            & 0.077$\pm$0.006      \\
\bottomrule
\end{tabular}
\label{tab:time test}
\end{minipage}
\begin{minipage}[t]{0.5\textwidth}
\makeatletter\def\@captype{table}
\small
\setcaptionwidth{0.9\textwidth}
\caption{\small{Sentiment classification accuracy on the IMDB test set of different BCLM discriminators.}}
\centering
\begin{tabular}{lc}
\toprule
Model   & Acc(\%) \\
\midrule
BCLM-GRU   & 88.4       \\
FUDGE(our implem.)   & 88.6       \\
Gemini-(w/o)KD            & 92.8      \\
\midrule
Gemini            & 89.2      \\
\bottomrule
\end{tabular}
\label{tab:acc}
\end{minipage}
\end{minipage} 

\subsection{Ablation Study}
\label{sec:ablation study}
To further verify the effectiveness of using the features extracted by GPT2 as input to discriminator, we designed a model BCLM-GRU, which differed from Gemini by using GRU~\cite{cho2014properties} as the backbone of the discriminator. 
The loss of the GRU was the cross-entropy loss from Equation~\eqref{equ:xe loss} in which we replaced $p_{\texttt{n}}(a|X_{1:i})$ with the output of the GRU. 
As shown in Table~\ref{tab:sentiment} and~\ref{tab:topic}, on both sentiment control and topic control, Gemini outperformed BCLM-GRU on the attribute relevance, perplexity and the excellent rate metrics. (Table~\ref{tab:sentiment} and~\ref{tab:topic})  The improvement could come from two possible sources: First, Gemini alleviated the mismatch between training and inference, and second, the discriminator of Gemini had better classification accuracy than the discriminator of BCLM-GRU. We report the sentiment classification accuracy of different BCLM discriminators on the IMDB test set in Table~\ref{tab:acc}. It is shown that the discriminator of Gemini obtained similar classification accuracy to the discriminator of BCLM-GRU, validating that the improvement came from the alleviating of the mismatch.

To validate the effectiveness of KD, we designed a
Gemini-(w/o)KD in which we removed the normal discriminator and the knowledge distillation loss. 
Gemini-(w/o)KD trained the faster discriminator with the cross-entropy loss from Equation~\eqref{equ:xe loss} in which we replaced $p_{\texttt{n}}(a|X_{1:i})$ with $p_{\texttt{f}}(a|X_{1:i-1},x_{i})$. 
Comparing Gemini with Gemini-(w/o)KD, we found that the knowledge distillation yielded great improvement on both sentiment control task and topic control task. (Table~\ref{tab:sentiment} and~\ref{tab:topic}) 
An interesting finding is that the performance of the discriminator of Gemini-(w/o)KD had better classification accuracy than that of Gemini (Table~\ref{tab:acc}) , showing that better classification accuracy of discriminator does not necessarily correspond to better performance of controllable language generation.

To analyse the effectiveness of the attribute-driven nucleus sampling, we designed Gemini-(w/o)AD, which differed from Gemini by adopting the nucleus sampling~\cite{topp} with a filter probability 0.7 as the decoding strategy. 
Comparing Gemini with Gemini-(w/o)AD, we found that the attribute-driven nucleus sampling had good performance on sentiment control. But on topic control, the improvement was not significant.
Moreover, it caused the decrease on the diversity metrics (i.e. Dist-1, Dist-2, Dist-3). (Table~\ref{tab:sentiment} and~\ref{tab:topic})
We think that both of the improvement on the controllability and the decrease on the diversity came from the strong restriction in the attribute-driven nucleus sampling.
Since Gemini could performs very well on the topic control task, the strong restriction of the attribute-driven nucleus sampling limited the performance of Gemini. 
To test the effect of attribute-driven nucleus sampling with other models, we applied it on FUDGE (Table~\ref{tab:fudge-ad}). On both of the tasks, the attribute-driven nucleus sampling significantly improved FUDGE.

\textbf{Stepwise Analysis of Discriminators.} Table~\ref{tab:stepwise} demonstrates the stepwise probabilities of positive attributes and sports topics. 
From the stepwise probabilities, we found that the stepwise probabilities assessed by Gemini flowed along with the conversion of the sentiment and topic in the sentences. For ``The food is awful but the service is nice'', the sentiment in the sentence converts from negative to positive, and the stepwise probability of positive attribute predicted by Gemini decreased firstly and then increased gradually. For ``The Italian athlete attended the World War in 1940'', the stepwise probability decreased after the word ``War'', which is reasonable. 
Compared with Gemini, BCLM-GRU and Gemini-(w/o)KD performed poorly on both of the sentences. They classified ``The food is awful but the service is nice'' as a very negative sentence, and classified ``The Italian athlete attended the World War in 1940'' as a sentence strongly related to the sports topic. Moreover, the stepwise probabilities assessed by BCLM-GRU and Gemini-(w/o)KD were usually extremely large or extremely small, which harmed their performance on controllable language generation.

\begin{table*}[t]
\small
\caption{\small{The stepwise probabilities assessed by different discriminators. The data in each step stands for the probability of the desired attribute for the sentence from the beginning to the current step.} }
\begin{center}
\begin{tabular}{l|ccccccccc}
\toprule
Positive Sentiment & The & food & is & awful & but & the & service & is & nice \\
\midrule
BCLM-GRU & - & 0.614 & 0.599 & 0.047 & 0.010& 0.005 & 0.003 & 0.002 & 0.002 \\
Gemini-(w/o)KD & - & 0.297 & 0.967 & 0.000 & 0.000 & 0.744 &0.000 &0.001 & 0.000 \\
Gemini & - & 0.538 & 0.556 & 0.171 & 0.371 & 0.480 & 0.456 &0.353 &0.581 \\
\midrule
Sports Topic& The & Italian & athlete & attended & the & World & War & in & 1940 \\
\midrule
BCLM-GRU & - & 0.077 & 0.964 & 0.999 & 1.000 & 1.000 & 1.000 & 1.000 & 0.999 \\
Gemini-(w/o)KD & - & 0.000 & 1.000 & 1.000 & 1.000 & 0.998 &0.999 &0.799 & 1.000 \\
Gemini & - & 0.113 & 0.777 & 0.599 & 0.846 & 0.798 &0.595 &0.268 & 0.240 \\
\bottomrule
\end{tabular}
\end{center}
\label{tab:stepwise}
\end{table*}
\section{Conclusion}
\label{sec:conclusion}
We propose Gemini, an efficient controllable language model that alleviates the mismatch between training and inference of BCLMs. Gemini reaches new state-of-the-art results on sentiment control and topic control. With strong controllability, the pre-trained language models will have more applications. However, although Gemini has great controllability, it performs poorly on diversity. There is still a long way to go to obtain very strong controllable language models.


\begin{thebibliography}{10}

\bibitem{GPT3}
Tom Brown, Benjamin Mann, Nick Ryder, Melanie Subbiah, Jared~D Kaplan, Prafulla
  Dhariwal, Arvind Neelakantan, Pranav Shyam, Girish Sastry, Amanda Askell,
  Sandhini Agarwal, Ariel Herbert-Voss, Gretchen Krueger, Tom Henighan, Rewon
  Child, Aditya Ramesh, Daniel Ziegler, Jeffrey Wu, Clemens Winter, Chris
  Hesse, Mark Chen, Eric Sigler, Mateusz Litwin, Scott Gray, Benjamin Chess,
  Jack Clark, Christopher Berner, Sam McCandlish, Alec Radford, Ilya Sutskever,
  and Dario Amodei.
\newblock Language models are few-shot learners.
\newblock In H.~Larochelle, M.~Ranzato, R.~Hadsell, M.~F. Balcan, and H.~Lin,
  editors, {\em Advances in Neural Information Processing Systems}, volume~33,
  pages 1877--1901. Curran Associates, Inc., 2020.

\bibitem{chan2020cocon}
Alvin Chan, Yew-Soon Ong, Bill Pung, Aston Zhang, and Jie Fu.
\newblock Cocon: A self-supervised approach for controlled text generation.
\newblock {\em arXiv preprint arXiv:2006.03535}, 2020.

\bibitem{chen2018temporal}
Hui Chen, Guiguang Ding, Sicheng Zhao, and Jungong Han.
\newblock Temporal-difference learning with sampling baseline for image
  captioning.
\newblock In {\em Proceedings of the AAAI Conference on Artificial
  Intelligence}, volume~32, 2018.

\bibitem{cho2014properties}
Kyunghyun Cho, Bart Van~Merri{\"e}nboer, Dzmitry Bahdanau, and Yoshua Bengio.
\newblock On the properties of neural machine translation: Encoder-decoder
  approaches.
\newblock {\em arXiv preprint arXiv:1409.1259}, 2014.

\bibitem{dathathri2019plug}
Sumanth Dathathri, Andrea Madotto, Janice Lan, Jane Hung, Eric Frank, Piero
  Molino, Jason Yosinski, and Rosanne Liu.
\newblock Plug and play language models: A simple approach to controlled text
  generation.
\newblock {\em arXiv preprint arXiv:1912.02164}, 2019.

\bibitem{bert}
Jacob Devlin, Ming{-}Wei Chang, Kenton Lee, and Kristina Toutanova.
\newblock {BERT:} pre-training of deep bidirectional transformers for language
  understanding.
\newblock In Jill Burstein, Christy Doran, and Thamar Solorio, editors, {\em
  Proceedings of the 2019 Conference of the North American Chapter of the
  Association for Computational Linguistics: Human Language Technologies,
  {NAACL-HLT} 2019, Minneapolis, MN, USA, June 2-7, 2019, Volume 1 (Long and
  Short Papers)}, pages 4171--4186. Association for Computational Linguistics,
  2019.

\bibitem{topk}
Angela Fan, Mike Lewis, and Yann Dauphin.
\newblock Hierarchical neural story generation.
\newblock {\em arXiv preprint arXiv:1805.04833}, 2018.

\bibitem{ficler-goldberg-2017-controlling}
Jessica Ficler and Yoav Goldberg.
\newblock Controlling linguistic style aspects in neural language generation.
\newblock In {\em Proceedings of the Workshop on Stylistic Variation}, pages
  94--104, Copenhagen, Denmark, September 2017. Association for Computational
  Linguistics.

\bibitem{distilling}
Geoffrey Hinton, Oriol Vinyals, and Jeffrey Dean.
\newblock Distilling the knowledge in a neural network.
\newblock In {\em NIPS Deep Learning and Representation Learning Workshop},
  2015.

\bibitem{hochreiter1997long}
Sepp Hochreiter and J{\"u}rgen Schmidhuber.
\newblock Long short-term memory.
\newblock {\em Neural computation}, 9(8):1735--1780, 1997.

\bibitem{topp}
Ari Holtzman, Jan Buys, Maxwell Forbes, and Yejin Choi.
\newblock The curious case of neural text degeneration.
\newblock {\em CoRR}, abs/1904.09751, 2019.

\bibitem{Hu2017Toward}
Zhiting Hu, Zichao Yang, Xiaodan Liang, Ruslan Salakhutdinov, and Eric~P. Xing.
\newblock Toward controlled generation of text.
\newblock In {\em Proceedings of the 34th International Conference on Machine
  Learning - Volume 70}, ICML'17, page 1587–1596. JMLR.org, 2017.

\bibitem{jiang2020can}
Zhengbao Jiang, Frank~F Xu, Jun Araki, and Graham Neubig.
\newblock How can we know what language models know?
\newblock {\em Transactions of the Association for Computational Linguistics},
  8:423--438, 2020.

\bibitem{keneshloo2019deep}
Yaser Keneshloo, Tian Shi, Naren Ramakrishnan, and Chandan~K Reddy.
\newblock Deep reinforcement learning for sequence-to-sequence models.
\newblock {\em IEEE transactions on neural networks and learning systems},
  31(7):2469--2489, 2019.

\bibitem{keskarCTRL2019}
Nitish~Shirish Keskar, Bryan McCann, Lav Varshney, Caiming Xiong, and Richard
  Socher.
\newblock {CTRL - A Conditional Transformer Language Model for Controllable
  Generation}.
\newblock {\em arXiv preprint arXiv:1909.05858}, 2019.

\bibitem{kikuchi-etal-2016-controlling}
Yuta Kikuchi, Graham Neubig, Ryohei Sasano, Hiroya Takamura, and Manabu
  Okumura.
\newblock Controlling output length in neural encoder-decoders.
\newblock In {\em Proceedings of the 2016 Conference on Empirical Methods in
  Natural Language Processing}, pages 1328--1338, Austin, Texas, November 2016.
  Association for Computational Linguistics.

\bibitem{kim-rush-2016-sequence}
Yoon Kim and Alexander~M. Rush.
\newblock Sequence-level knowledge distillation.
\newblock In {\em Proceedings of the 2016 Conference on Empirical Methods in
  Natural Language Processing}, pages 1317--1327, Austin, Texas, November 2016.
  Association for Computational Linguistics.

\bibitem{krause-etal-2021-gedi-generative}
Ben Krause, Akhilesh~Deepak Gotmare, Bryan McCann, Nitish~Shirish Keskar,
  Shafiq Joty, Richard Socher, and Nazneen~Fatema Rajani.
\newblock {G}e{D}i: Generative discriminator guided sequence generation.
\newblock In {\em Findings of the Association for Computational Linguistics:
  EMNLP 2021}, pages 4929--4952, Punta Cana, Dominican Republic, November 2021.
  Association for Computational Linguistics.

\bibitem{li-etal-2016-diversity}
Jiwei Li, Michel Galley, Chris Brockett, Jianfeng Gao, and Bill Dolan.
\newblock A diversity-promoting objective function for neural conversation
  models.
\newblock In {\em Proceedings of the 2016 Conference of the North {A}merican
  Chapter of the Association for Computational Linguistics: Human Language
  Technologies}, pages 110--119, San Diego, California, June 2016. Association
  for Computational Linguistics.

\bibitem{Li2021PrefixTuningOC}
Xiang~Lisa Li and Percy Liang.
\newblock Prefix-tuning: Optimizing continuous prompts for generation.
\newblock {\em Proceedings of the 59th Annual Meeting of the Association for
  Computational Linguistics and the 11th International Joint Conference on
  Natural Language Processing (Volume 1: Long Papers)}, abs/2101.00190, 2021.

\bibitem{liu2021vocab}
Han Liu, Shifeng Zhang, Ke~Lin, Jing Wen, Jianmin Li, and Xiaolin Hu.
\newblock Vocabulary-wide credit assignment for training image captioning
  models.
\newblock {\em IEEE Transactions on Image Processing}, 30:2450--2460, 2021.

\bibitem{liu2021pre}
Pengfei Liu, Weizhe Yuan, Jinlan Fu, Zhengbao Jiang, Hiroaki Hayashi, and
  Graham Neubig.
\newblock Pre-train, prompt, and predict: A systematic survey of prompting
  methods in natural language processing.
\newblock {\em arXiv preprint arXiv:2107.13586}, 2021.

\bibitem{loshchilov2017decoupled}
Ilya Loshchilov and Frank Hutter.
\newblock Decoupled weight decay regularization.
\newblock {\em arXiv preprint arXiv:1711.05101}, 2017.

\bibitem{maas-etal-2011-imdb}
Andrew~L. Maas, Raymond~E. Daly, Peter~T. Pham, Dan Huang, Andrew~Y. Ng, and
  Christopher Potts.
\newblock Learning word vectors for sentiment analysis.
\newblock In {\em Proceedings of the 49th Annual Meeting of the Association for
  Computational Linguistics: Human Language Technologies}, pages 142--150,
  Portland, Oregon, USA, June 2011. Association for Computational Linguistics.

\bibitem{papineni2002bleu}
Kishore Papineni, Salim Roukos, Todd Ward, and Wei-Jing Zhu.
\newblock Bleu: a method for automatic evaluation of machine translation.
\newblock In {\em Proceedings of the 40th annual meeting on Association for
  Computational Linguistics}, pages 311--318. Association for Computational
  Linguistics, 2002.

\bibitem{qian2022prefixtuning}
Jing Qian, Li~Dong, Yelong Shen, Furu Wei, and Weizhu Chen.
\newblock Controllable natural language generation with contrastive prefixes.
\newblock {\em arXiv preprint arXiv:2202.13257}, 2022.

\bibitem{radford2017learning}
Alec Radford, Rafal Jozefowicz, and Ilya Sutskever.
\newblock Learning to generate reviews and discovering sentiment.
\newblock {\em arXiv preprint arXiv:1704.01444}, 2017.

\bibitem{radford2018improving}
Alec Radford, Karthik Narasimhan, Tim Salimans, and Ilya Sutskever.
\newblock Improving language understanding by generative pre-training.
\newblock 2018.

\bibitem{radford2019language}
Alec Radford, Jeffrey Wu, Rewon Child, David Luan, Dario Amodei, Ilya
  Sutskever, et~al.
\newblock Language models are unsupervised multitask learners.
\newblock {\em OpenAI blog}, 1(8):9, 2019.

\bibitem{ranzato2015sequence}
Marc'Aurelio Ranzato, Sumit Chopra, Michael Auli, and Wojciech Zaremba.
\newblock Sequence level training with recurrent neural networks.
\newblock In {\em 4th International Conference on Learning Representations},
  2016.

\bibitem{rennie2017self}
Steven~J Rennie, Etienne Marcheret, Youssef Mroueh, Jerret Ross, and Vaibhava
  Goel.
\newblock Self-critical sequence training for image captioning.
\newblock In {\em Proceedings of the IEEE Conference on Computer Vision and
  Pattern Recognition}, pages 7008--7024, 2017.

\bibitem{Shin2020Auto}
Taylor Shin, Yasaman Razeghi, Robert IV, Eric Wallace, and Sameer Singh.
\newblock Autoprompt: Eliciting knowledge from language models with
  automatically generated prompts.
\newblock pages 4222--4235, 01 2020.

\bibitem{tesauro1995temporal}
Gerald Tesauro et~al.
\newblock Temporal difference learning and td-gammon.
\newblock {\em Communications of the ACM}, 38(3):58--68, 1995.

\bibitem{vedantam2015cider}
Ramakrishna Vedantam, C~Lawrence~Zitnick, and Devi Parikh.
\newblock Cider: Consensus-based image description evaluation.
\newblock In {\em Proceedings of the IEEE Conference on Computer Vision and
  Pattern Recognition}, pages 4566--4575, 2015.

\bibitem{wolf2020transformers}
Thomas Wolf, Julien Chaumond, Lysandre Debut, Victor Sanh, Clement Delangue,
  Anthony Moi, Pierric Cistac, Morgan Funtowicz, Joe Davison, Sam Shleifer,
  et~al.
\newblock Transformers: State-of-the-art natural language processing.
\newblock In {\em Proceedings of the 2020 Conference on Empirical Methods in
  Natural Language Processing: System Demonstrations}, pages 38--45, 2020.

\bibitem{yang2021fudge}
Kevin Yang and Dan Klein.
\newblock Fudge: Controlled text generation with future discriminators.
\newblock {\em arXiv preprint arXiv:2104.05218}, 2021.

\bibitem{yang2019xlnet}
Zhilin Yang, Zihang Dai, Yiming Yang, Jaime Carbonell, Russ~R Salakhutdinov,
  and Quoc~V Le.
\newblock Xlnet: Generalized autoregressive pretraining for language
  understanding.
\newblock {\em Advances in neural information processing systems}, 32, 2019.

\bibitem{yu2021attribute}
Dian Yu, Zhou Yu, and Kenji Sagae.
\newblock Attribute alignment: Controlling text generation from pre-trained
  language models.
\newblock {\em arXiv preprint arXiv:2103.11070}, 2021.

\bibitem{SeqGAN}
Lantao Yu, Weinan Zhang, Jun Wang, and Yong Yu.
\newblock Seqgan: Sequence generative adversarial nets with policy gradient.
\newblock In {\em Proceedings of the Thirty-First AAAI Conference on Artificial
  Intelligence}, AAAI'17, page 2852–2858. AAAI Press, 2017.

\bibitem{zhang2017actor}
Li~Zhang, Flood Sung, Feng Liu, Tao Xiang, Shaogang Gong, Yongxin Yang, and
  Timothy~M Hospedales.
\newblock Actor-critic sequence training for image captioning.
\newblock In {\em NIPS Workshop on Visually-Grounded Interaction and Language},
  2017.

\bibitem{agnews}
Xiang Zhang, Junbo Zhao, and Yann LeCun.
\newblock Character-level convolutional networks for text classification.
\newblock In {\em Proceedings of the 28th International Conference on Neural
  Information Processing Systems - Volume 1}, NIPS'15, page 649–657,
  Cambridge, MA, USA, 2015. MIT Press.

\end{thebibliography}

\newpage

\appendix
\section{Appendix}
\subsection{Prefixes used in experiments for sentence generation}
\label{sec:prefixes}
In both automatic and human evaluation, we used the same 15 prefixes for sentiment control and 20 prefixes for topic control from PPLM.

\textbf{Prefixes for sentiment control:} ``Once upon a time'', ``The
book'', ``The chicken'', ``The city'', ``The country'', ``The horse'', ``The lake'', ``The last time'',``The movie'', ``The painting'', ``The pizza'', ``The potato'', ``The president of the country'', ``The road'', ``The year is 1910.''.

\textbf{Prefixes for topic control:} ``In summary,'', ``This essay discusses'', ``Views on'', ``The connection'', ``Foundational to this is'', ``To review,'', ``In brief,'',
``An illustration of'', ``Furthermore,'', ``The central theme'', ``To conclude,'', ``The key aspect'', ``Prior to this'', ``Emphasised are'',
``To summarise,'', ``The relationship'', ``More importantly,'', ``It has been shown'', ``The issue focused on'', ``In this essay''.

\subsection{More details for the human evaluation experiments}
\label{sec:details human}
\begin{figure*}[htbp]
  \centering
  \includegraphics[width=\textwidth]{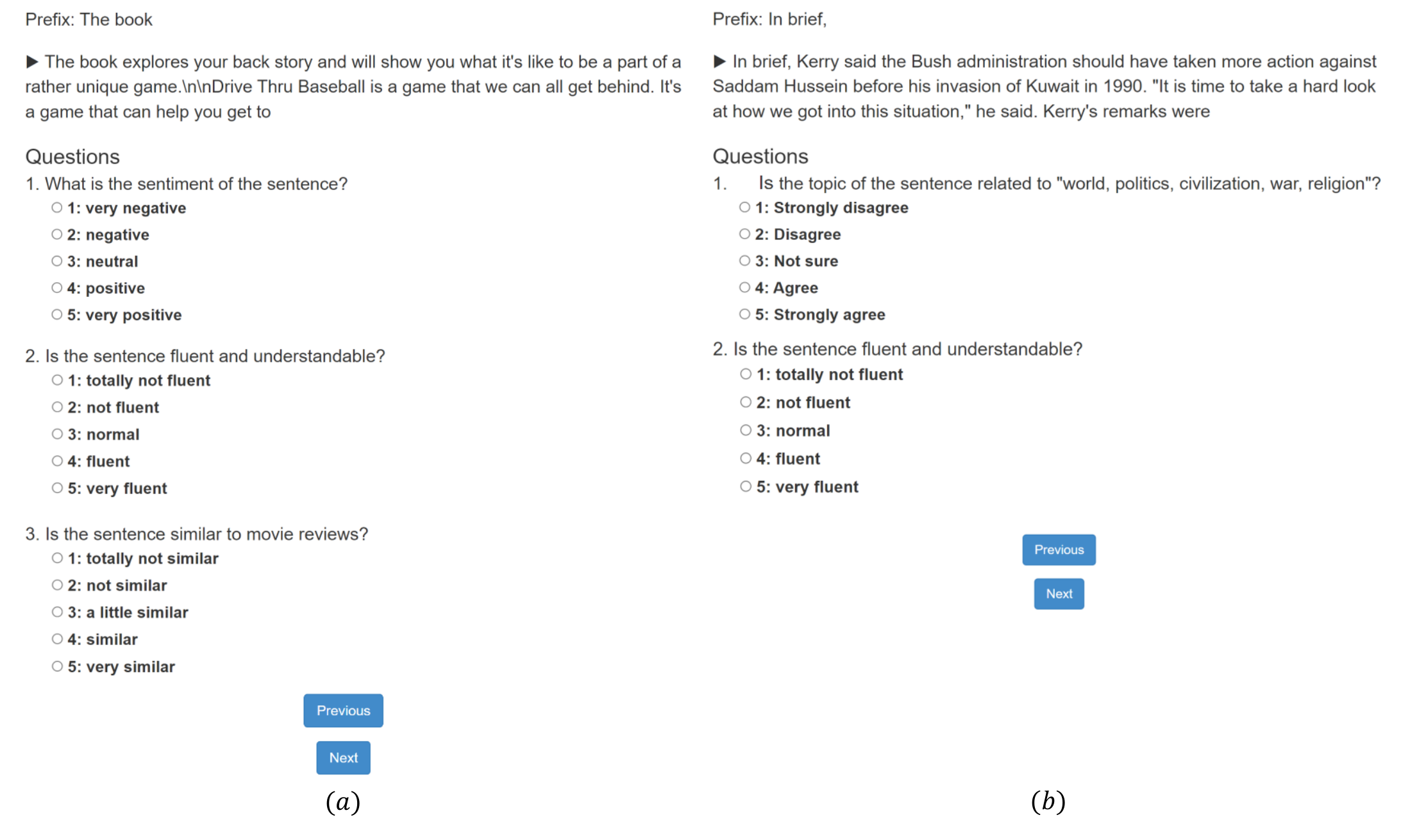}
  \caption{\small{The interface of the webpage for sentiment control (a) and topic control (b).}}
\label{fig:website}
\end{figure*}

In human evaluation of both sentiment control and topic control, the test questions were presented to annotators in a webpage. 
The interfaces of the webpage for sentiment control and topic control are demonstrated in Figure~\ref{fig:website}.
For each generated sample, the webpage demonstrated the prefix, the generated sample and the questions for the task. For sentiment control, there were three questions per sample: the sentiment of the sentence, the linguistic quality of the sentence, and the similarity between the sentence and movie reviews. For topic control, there were two questions per sample: the relevance between the sentence and the desired topic, and the linguistic quality of the sentence. 
The desired topics were expressed via its keywords. The following four sets of keywords represented four different topics: (1) world, politics, civilization and war, (2) sports, (3) business, economy and finance, (4) science and technology.
All questions were scored on 1-5 Likert scale. 

In experiments, we randomized the samples generated by different models. Each annotator for sentiment/topic control were asked to evaluate 60/80 samples, respectively. On average, it took an annotator about 40s to evaluate a sample for either task.

\subsection{Model structure of normal discriminator and faster discriminator}
In this section we introduce the detailed structure of the normal discriminator and the faster discriminator in Gemini.
Let $h_1,...,h_T$ denote the last hidden state of $G$ with $X_{1:T}$ as the input. For each $i$, the dimension of $h_{i}$ is $d_h$.
The normal discriminator $M_{\texttt{n}}$ receives $h_{i}$ as the input.
The structure of $M_{\texttt{n}}$ is a two-layer neural network:
\begin{equation}
\begin{aligned}
&g_{i}=\texttt{ReLU}(W_{g}h_{i}+b_{g}), \\
&M_{\texttt{n}}(h_{i})=W_{0}g_{i},
\label{equ:normal structure}
\end{aligned}
\end{equation}
where $W_{g}$ is a $d_h \times d_h$ matrix, $b_{g}$ is a $d_h$ dimensional vector, $W_0$ is a $d_h$ dimensional row vector and ReLU is the rectified linear activation function.

Let $e_{w}$ denote the embedding of token $w$. The dimension of $e_{w}$ is $d_e$.
The faster discriminator $M_{\texttt{f}}$ receives $h_{i-1}$ and $e_{w}$ as the input.
The structure of $M_{\texttt{f}}$ is a gated neural network:
\begin{equation}
\begin{aligned}
&g_{i-1}=W_{g}h_{i-1}+b_{g}, \\
&r_{i}=\sigma(W_{1r}e_{w}+W_{2r}g_{i-1}+b_{r}), \\
&z_{i}=\sigma(W_{1z}e_{w}+W_{2z}g_{i-1}+b_{z}), \\
&n_{i}=\texttt{Tanh}(W_{1n}e_{w}+b_{1n}+r_{i}(W_{2n}g_{i-1}+b_{2n})),\\
&o_{i}=\texttt{ReLU}((1-z_{i})n_{i}+z_{i}g_{i-1}),\\
&M_{\texttt{f}}(h_{i-1},e_{w})=W_0o_{i},
\label{equ:fast structure}
\end{aligned}
\end{equation}
where $W_{1r}$,$W_{1z}$,$W_{1n}$ are $d_h \times d_e$ matrices, $W_{2r}$,$W_{2z}$,$W_{2n}$ are $d_h \times d_h$ matrices, $b_r$, $b_z$, $b_{1n}$ and $b_{2n}$ are $d_h$ dimensional vectors and Tanh is the hyperbolic tangent function.
Specifically, we reuse $W_g$, $b_g$ and $W_0$ in Equation~\eqref{equ:normal structure} to reduce the parameters of our model.

\subsection{More generated samples}
We show randomly selected samples with sentiment control and topic control generated by different models in Tables~\ref{tab:sentiment samples} and~\ref{tab:topic samples}, respectively.

\begin{table*}[htbp]
\small
\caption{\small{Generated sentences with sentiment control. Words underscored are the prefixes. Words that signify corpus overfitting are highlighted in blue. Words that signify the controlled effect are highlighted in red.} }
\begin{center}
\resizebox{\textwidth}{95mm}{
\begin{tabular}{l|l}
\toprule
\tabincell{l}{Model\\\#Sentiment} & Generated Sentence\\
\midrule
\tabincell{l}{Gemini\\\#Positive} & \tabincell{l}{\uline{The book} is \textcolor{red}{beautifully} illustrated and gives a \textcolor{red}{great} account of the history of the American Indian,\\ and the \textcolor{red}{many varied and interesting} people who have been the subjects of his own \textcolor{red}{wonderful} narrative. } \\
\midrule
\tabincell{l}{Gemini\\\#Negative} & \tabincell{l}{\uline{The book} \textcolor{red}{isn't much of} an improvement, and the only thing that can make it \textcolor{red}{worse} is \textcolor{red}{a lack of} \\any real sense of humor. The writing is \textcolor{red}{dreadful}, and the characters are \textcolor{red}{bad, horrible, terrible.}} \\
\midrule
\tabincell{l}{GPT2-ranked\\\#Positive} & \tabincell{l}{\uline{The book} will tell you all about the world of The Legend of Zelda, including its history, characters, \\enemies and bosses. It will also show you how to play this game. This is a \textcolor{red}{great} book for anyone}\\
\midrule
\tabincell{l}{GPT2-ranked\\\#Negative} & \tabincell{l}{\uline{The book} also features an essay by Dr. Stephen Siegel, the president of the American Psychiatric \\Association, and a foreword from Dr. Robert Lustig, a leading psychiatrist who has been \textcolor{red}{critical}} \\
\midrule
\tabincell{l}{GPT2-tune\\\#Positive} & \tabincell{l}{\uline{The book} was a very \textcolor{red}{enjoyable} read, and I think the \textcolor{blue}{film} is \textcolor{red}{much better}. But it has many flaws that \\detract from its value as a \textcolor{blue}{movie}.<br /><br />It begins with the story of an orphaned boy who lives} \\
\midrule
\tabincell{l}{GPT2-tune\\\#Negative} & \tabincell{l}{\uline{The book} itself was \textcolor{red}{a bit dry and dull}. The story is well done, however the script was \textcolor{red}{very weak} in \\places. It seems like an attempt to get attention from \textcolor{red}{critics} for a \textcolor{blue}{film} that may or may not ever be made.}\\
\midrule
\tabincell{l}{PPLM\\\#Positive} & \tabincell{l}{\uline{The book}$\backslash$n$\backslash$nThe book is a collection of the \textcolor{red}{best} books of the \textcolor{red}{best} books of the \textcolor{red}{best} of the \textcolor{red}{best} \\of the \textcolor{red}{best} (or "the \textcolor{red}{best} of the \textcolor{red}{best} of the \textcolor{red}{best}" if you prefer the more technical meaning).} \\
\midrule
\tabincell{l}{PPLM\\\#Negative} & \tabincell{l}{\uline{The book} is out, the game is in, and it's time for an update on what's going on with this. I. I'm \textcolor{red}{not} \\so much for that. \textcolor{red}{I'm not.}$\backslash$n$\backslash$nI'm \textcolor{red}{a little, a little, a little,}}\\
\midrule
\tabincell{l}{GeDi\\\#Positive} & \tabincell{l}{\uline{The book} traces how these communities gathered to face prejudice and integration as they reacquainted \\themselves with one another during purges in Poland in the 1940s.$\backslash$n$\backslash$nElsewhere around the world} \\
\midrule
\tabincell{l}{GeDi\\\#Negative} & \tabincell{l}{\uline{The book} \textcolor{red}{sucked} again. Stephen says it's only.500 on his blog, but I had to crawl through ten-thousand\\ pages of text on a beautiful new iPad before getting into the meat and potatoes of Uncharted 4: A Thief's}\\
\midrule
\tabincell{l}{FUDGE\\\#Positive} & \tabincell{l}{\uline{The book} is based on interviews with more than 100 people about their experience working at a number \\of different companies, from the \textcolor{red}{likes} of Nike and Microsoft to Amazon and Google. } \\
\midrule
\tabincell{l}{FUDGE\\\#Negative} & \tabincell{l}{\uline{The book} is also the only one to have been published in both English and Chinese. This book contains \\the most complete collection of the writings on Buddhism in all languages. The author has spent much}\\
\midrule
\midrule
\tabincell{l}{Gemini\\\#Positive} & \tabincell{l}{\uline{Once upon a time}, our family \textcolor{red}{enjoyed} a \textcolor{red}{vibrant} and \textcolor{red}{dynamic} community with a \textcolor{red}{strong} sense of community and \\having a life that was \textcolor{red}{truly} yours. The lives we had were \textcolor{red}{extremely full} and gave us a \textcolor{red}{unique and wonderful}} \\
\midrule
\tabincell{l}{Gemini\\\#Negative} & \tabincell{l}{\uline{Once upon a time}, Once upon a time, when everything \textcolor{red}{wasn't going too well}, they tried to cook something \\better and somehow seemed to have made a \textcolor{red}{shitty mess} of the \textcolor{red}{bad} stuff. Anybody with a decent amount of} \\
\midrule
\tabincell{l}{GPT2-ranked\\\#Positive} & \tabincell{l}{\uline{Once upon a time}, the first person to walk through that door was my grandfather. He was in his 80s \\when he came here, and it took him almost four hours to get there because we had an enormous}\\
\midrule
\tabincell{l}{GPT2-ranked\\\#Negative} & \tabincell{l}{\uline{Once upon a time}, there was a man who had the ability to travel through space and time. He could teleport \\anywhere he wished at any moment without being noticed. But one day, his ability \textcolor{red}{got lost} in the ether,} \\
\midrule
\tabincell{l}{GPT2-tune\\\#Positive} & \tabincell{l}{\uline{Once upon a time}, I was \textcolor{red}{in love} with \textcolor{blue}{the film "The Manchurian Candidate"}. The story is very similar to \\that of this \textcolor{blue}{movie} and the \textcolor{blue}{characters} are pretty much identical. But one thing that separates it from its} \\
\midrule
\tabincell{l}{GPT2-tune\\\#Negative} & \tabincell{l}{\uline{Once upon a time}, I would have liked to see the sequel. I loved the first \textcolor{blue}{film} and thought that it was very \\good, but now I think that this \textcolor{blue}{movie} is \textcolor{red}{much worse} than the first. It's not even close to what the original}\\
\midrule
\tabincell{l}{PPLM\\\#Positive} & \tabincell{l}{\uline{Once upon a time}, a man, a \textcolor{red}{wise} man, a \textcolor{red}{noble}, the \textcolor{red}{best} of a \textcolor{red}{wise} man, said to his friends, "I am a very \\\textcolor{red}{great} man in the world. And I know a \textcolor{red}{great} mystery which \textcolor{red}{a great many great} men do not know.} \\
\midrule
\tabincell{l}{PPLM\\\#Negative} & \tabincell{l}{\uline{Once upon a time}, in the distant past, there was a \textcolor{red}{world-ending monster}, an \textcolor{red}{enormous, evil, ugly, nasty,} \\\textcolor{red}{disgusting, boring, boring, boring, boring, boring, boring, boring, boring, boring,}}\\
\midrule
\tabincell{l}{GeDi\\\#Positive} & \tabincell{l}{\uline{Once upon a time}…is a \textcolor{red}{beautiful musical triumph of bold harmonies}, \textcolor{red}{exceptional} lyrical expression, \\\textcolor{red}{beautifully} delivered words and \textcolor{red}{unforgettable} music. Indeed the poetry itself is an embodiment} \\
\midrule
\tabincell{l}{GeDi\\\#Negative} & \tabincell{l}{\uline{Once upon a time}, there connected to save data on Fitbit Wristbands. \textcolor{red}{Unfortunately} this feature is meant \\to record your heart rate and then compare it to a threshold value that slowly ramps up the burn after }\\
\midrule
\tabincell{l}{FUDGE\\\#Positive} & \tabincell{l}{\uline{Once upon a time}, the \textcolor{red}{great} king of the elves, Vilya and his wife Elvira, were living in the \textcolor{red}{beautiful} \\city of Alkenstar, on the world of Erenheim. There Vilya was \textcolor{red}{married to the powerful}} \\
\midrule
\tabincell{l}{FUDGE\\\#Negative} & \tabincell{l}{\uline{Once upon a time}, I had a small apartment in New York City, and it had a huge bathroom. I thought about \\putting the shower in there, but I didn't know how big the tub was, because I never saw the tub from the}\\
\bottomrule
\end{tabular}}
\end{center}
\label{tab:sentiment samples}
\end{table*}
\begin{table*}[htbp]
\small
\caption{\small{Generated sentences with topic control. Words underscored are the prefixes. Words that signify the controlled effect are highlighted in red.} }
\begin{center}
\resizebox{\textwidth}{90mm}{
\begin{tabular}{l|l}
\toprule
\tabincell{l}{Model\\\#Topic} & Generated Sentence\\
\midrule
\tabincell{l}{Gemini\\\#World} & \tabincell{l}{\uline{The issue focused on} \textcolor{red}{President Barack Obama's election} in 2008 and his \textcolor{red}{political} victory in 2012, in which \\\textcolor{red}{Republican presidential nominee} Mitt Romney's national conservative movement was the \textcolor{red}{party's opposition}. \\\textcolor{red}{Democrats} contested \textcolor{red}{Republicans'} electoral victories in a two-party system in a presidential} \\
\midrule
\tabincell{l}{Gemini\\\#Sports} & \tabincell{l}{\uline{The issue focused on} \textcolor{red}{baseball's} new international \textcolor{red}{player signing program}. The \textcolor{red}{league}, which this \textcolor{red}{offseason} \\designated two \textcolor{red}{Canadian players} for the playoffs, said the one \textcolor{red}{player} who was signed from the \textcolor{red}{CFL} and the \\two \textcolor{red}{players} signed from the \textcolor{red}{American League} had no right to play} \\
\midrule
\tabincell{l}{Gemini\\\#Business} & \tabincell{l}{\uline{The issue focused on} \textcolor{red}{tax rates in the company's taxation and the tax rate paid by the retailer}, which raised \\the biggest \textcolor{red}{fiscal cost} for the \textcolor{red}{cash-strapped retailer}. $\backslash$n$\backslash$nThe \textcolor{red}{tax rate} the retailer paid on \textcolor{red}{earnings} in the \\first three months of}\\
\midrule
\tabincell{l}{Gemini\\\#Science} & \tabincell{l}{\uline{The issue focused on} \textcolor{red}{software} that can detect \textcolor{red}{malware and online threats and a JavaScript framework} that \\can be used to identify \textcolor{red}{web sites} that are malicious or that could be the source of a \textcolor{red}{Web-based infection}.\\$\backslash$n$\backslash$nThe \textcolor{red}{cyber security researchers}, working with \textcolor{red}{technology}} \\
\midrule
\midrule
\tabincell{l}{GPT2-tune\\\#World} & \tabincell{l}{\uline{The issue focused on} the number of \textcolor{red}{casualties}, but \textcolor{red}{a spokesman for Hamas}, which controls \textcolor{red}{Gaza}, called for \\calm in response to \textcolor{red}{Israel's actions}. The \textcolor{red}{Palestinian leader}, Yasser Arafat "is still alive and his position \\is unchanged"} \\
\midrule
\tabincell{l}{GPT2-tune\\\#Sports} & \tabincell{l}{\uline{The issue focused on} whether or not the \textcolor{red}{Red Sox} could make a run at another title with their The \textcolor{red}{pitching} staff. \\And, according to sources close to Boston, that was what the \textcolor{red}{team's} new front office was looking for in \\their top two} \\
\midrule
\tabincell{l}{GPT2-tune\\\#Business} & \tabincell{l}{\uline{The issue focused on} a recent federal decision to \textcolor{red}{raise taxes} for some \textcolor{red}{high-income earners} and how it might \\affect the ability of \textcolor{red}{companies} to make \textcolor{red}{investments} in their own \textcolor{red}{businesses}.  "There is no single \textcolor{red}{tax rate} that \\should be applied to all Americans}\\
\midrule
\tabincell{l}{GPT2-tune\\\#Science} & \tabincell{l}{\uline{The issue focused on} the size of the \textcolor{red}{new system}, which is scheduled to go into mass production by late 2006. \\It also includes a host of \textcolor{red}{software features} designed to improve performance and reduce costs. <FONT face=\\"verdana,} \\
\midrule
\midrule
\tabincell{l}{GeDi\\\#World} & \tabincell{l}{\uline{The issue focused on} \textcolor{red}{Nicaragua's peace process} with the \textcolor{red}{Revolutionary Armed Forces of Colombia (FARC)}; \\its electoral participation; and human rights. Moreover, Ambassador Carr called on private actors responsible \\for violence in places like \textcolor{red}{Buenaventura} were to name} \\
\midrule
\tabincell{l}{GeDi\\\#Sports} & \tabincell{l}{\uline{The issue focused on} Rodriguez not throwing his \textcolor{red}{second pitch} (on what appeared to be a \textcolor{red}{fastball} against \\Perez) in a \textcolor{red}{slow-pitch} count of seven. Another item nixed: satire by Sheldon Dreyer arguing that because video \\from MLB Advanced} \\
\midrule
\tabincell{l}{GeDi\\\#Business} & \tabincell{l}{\uline{The issue focused on} Costco Wholesale Inc's \textcolor{red}{debt}.... \textcolor{red}{Financials} were shed as Welch was quoted saying he \\wanted to remain long in real estate with years to restructure it such that shareholders share in \textcolor{red}{annual payouts} \\if they made well and}\\
\midrule
\tabincell{l}{GeDi\\\#Science} & \tabincell{l}{\uline{The issue focused on} \textcolor{red}{Linux kernel version}, as well as \textcolor{red}{network stack statistics} (how many different \textcolor{red}{host} \\\textcolor{red}{connections} are being made). "I've drafted mixed thoughts for various types of \textcolor{red}{attack models}", concedes \\Adam Wexler. The answer is to make you} \\
\midrule
\midrule
\tabincell{l}{FUDGE\\\#World} & \tabincell{l}{\uline{The issue focused on} \textcolor{red}{President Barack Obama's} pledge to deport all children who came here illegally as \\children, saying "no child should have to face deportation," and the \textcolor{red}{GOP's} refusal to accept any of that border \\wall funding.$\backslash$n$\backslash$n\textcolor{red}{Republican} senators were divided} \\
\midrule
\tabincell{l}{FUDGE\\\#Sports} & \tabincell{l}{\uline{The issue focused on} the role that \textcolor{red}{players played in the game}. \textcolor{red}{Players} were allowed to "call out" \textcolor{red}{referees} or \\\textcolor{red}{referees} should not \textcolor{red}{referee} their \textcolor{red}{team}, and \textcolor{red}{referees} responded by calling \textcolor{red}{players} out.$\backslash$n$\backslash$n\textcolor{red}{Fans} reacted to the \\incident by booing \textcolor{red}{referees} and} \\
\midrule
\tabincell{l}{FUDGE\\\#Business} & \tabincell{l}{\uline{The issue focused on} \textcolor{red}{financial incentives for banks} to \textcolor{red}{borrow money from taxpayers}. Federal Reserve chair \\Janet Yellen said the \textcolor{red}{bank's mortgage-backed securities} are risky investments, and that \textcolor{red}{banks} must \textcolor{red}{pay} more \\than \textcolor{red}{banks} were \textcolor{red}{earning} before the \textcolor{red}{housing crisis}. Treasury Secretary}\\
\midrule
\tabincell{l}{FUDGE\\\#Science} & \tabincell{l}{\uline{The issue focused on} the use of \textcolor{red}{software to monitor and hack into phones}, \textcolor{red}{computer systems} that allow users \\to encrypt their communications or remotely control \textcolor{red}{devices such as cameras and TVs}. \textcolor{red}{Apple} has since \\updated the \textcolor{red}{iOS operating system} and introduced a new \textcolor{red}{encryption option in iOS}} \\
\bottomrule
\end{tabular}}
\end{center}
\label{tab:topic samples}
\end{table*}

\subsection{More discussion on decoding strategy}
We tested the effectiveness of the attribute-driven nucleus sampling on another BCLM method FUDGE. The decoding strategy in the original paper of FUDGE was the top-k strategy with $k=10$. The FUDGE model with the attribute-driven nucleus sampling is called ``FUDGE-AD''. The results are shown in Table~\ref{tab:fudge-ad}. 
On sentiment control, the attribute-driven nucleus sampling significantly improved attribute relevance and excellent rate, but it had worse perplexity. And on topic control, the attribute-driven nucleus sampling had better performance on attribute relevance, perplexity and excellent rate.
Specifically, compared to the top-k decoding strategy, the attribute-driven nucleus sampling had similar diversity on both tasks.

\begin{table*}[t]
\small
\caption{\small{Automatic evaluation of FUDGE and FUDGE-AD on sentiment control and topic control.} }
\begin{center}
\begin{tabular}{l|ccccccc}
\toprule
Model &AR(\%)$\uparrow$ & PPL$\downarrow$ & ER(\%)$\uparrow$ & Dist-1$\uparrow$ & Dist-2$\uparrow$ & Dist-3$\uparrow$ & CR(\%)$\downarrow$\\
\midrule
\multicolumn{8}{l}{\textit{Sentiment Control}} \\
\midrule
FUDGE &  70.9 & \textbf{11.06} & 18.1 & 0.39 & 0.79 & 0.89 & \textbf{1.19} \\
FUDGE-AD & \textbf{93.6} & 32.83 & \textbf{35.2} & \textbf{0.44} & \textbf{0.82} & \textbf{0.90}  &5.24  \\
\midrule
\multicolumn{8}{l}{\textit{Topic Control}} \\
\midrule
FUDGE & 85.1 & 14.52 & 77.8 & \textbf{0.38} &\textbf{0.79} & 0.88 & \textbf{7.53} \\
FUDGE-AD & \textbf{96.9} & \textbf{12.43} & \textbf{95.2} & \textbf{0.38} & 0.78 & \textbf{0.90}  &12.74  \\
\bottomrule
\end{tabular}
\end{center}
\label{tab:fudge-ad}
\end{table*}

\begin{figure*}[htbp]
  \centering
  \includegraphics[width=\textwidth]{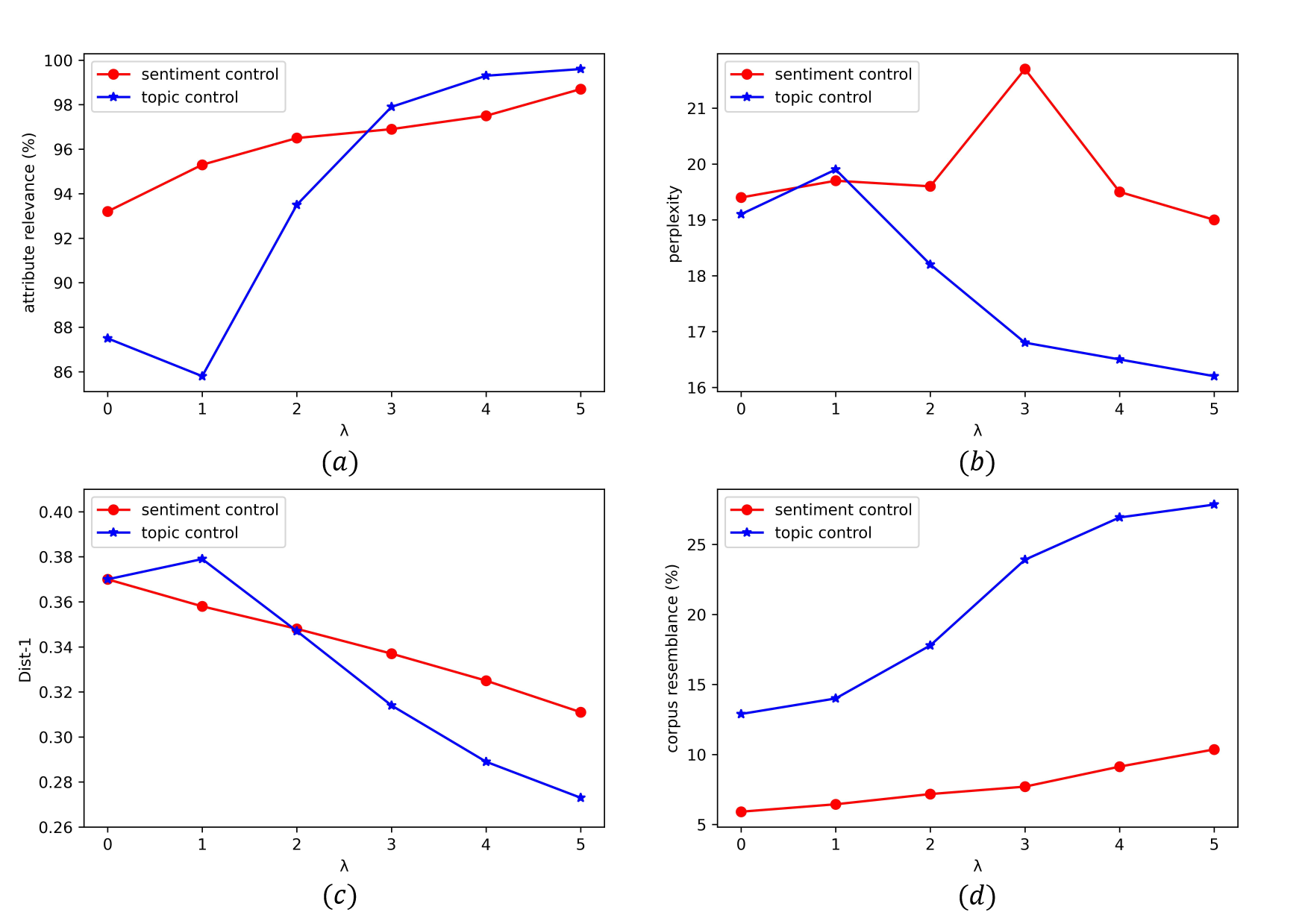}
  \caption{\small{Automatic evluation of Gemini with different $\lambda$ on attribute relevance (a), perplexity (b), Dist-1 (c), and corpus resemblance (d).}}
\label{fig:lambda}
\end{figure*}

We investigated the effectiveness of the hyper-parameter $\lambda$ in the weighted decoding (Equation~\eqref{equ:weighted decoding}). Figure~\ref{fig:lambda} demonstrates the automatic evaluation of Gemini with different $\lambda$. It is shown that on both sentiment control and topic control, with the increase of $\lambda$, the attribute relevance (Figure~\ref{fig:lambda}(a)) and the corpus resemblance (Figure~\ref{fig:lambda}(d)) increased, while the diversity (Figure~\ref{fig:lambda}(c)) decreased.
With the increase of $\lambda$,
the perplexity for sentiment control was not significantly affected, while the perplexity for topic control decreased (Figure~\ref{fig:lambda}(b)). 

\subsection{Multi-attributes control experiments}
In the main text of our paper, we have discussed the experiments for single attribute control tasks. In this section, we discuss the performance of Gemini on multi-attributes control.
To condition on multiple attributes $a_1,a_2,...,a_k$, we need to model the conditional probability 
\begin{equation}
\begin{aligned}
p(X|a_1,a_2,...,a_k)=\prod_{i=1}^{T}p(x_{i}|X_{1:i-1},a_1,a_2,...,a_k).
\label{equ:multi-attribute p}
\end{aligned}
\end{equation}
If the attributes are mutually independent, according to the Bayes Rule, 
\begin{equation}
\begin{aligned}
p(x_{i}|X_{1:i-1},a_1,a_2,...,a_k)&\propto p(a_1,a_2,...,a_k|X_{1:i})p(x_{i}|X_{1:i-1})\\
&\propto \prod_{j=1}^{k}p(a_j|X_{1:i})p(x_{i}|X_{1:i-1}).
\label{equ:bayes multi-attribute}
\end{aligned}
\end{equation}
The likelihood of multiple attributes can be modeled with the likelihood of each independent attribute. 
We experimented on sentiment-topic control (e.g. positive + world and negative + sports). We used the Gemini models trained on single-attribute control (sentiment control and topic control) to model $p(a_j|X_{1:i})$ in Equation~\eqref{equ:bayes multi-attribute}.

The baseline model we used in the experiment is named ``GPT2-tune-MA''. In Section~\ref{sec:baselines}, we used prefix-tuning in training the GPT2-tune models. Prefix-tuning introduced a sequence of trainable continuous vectors to input, named ``prefix vectors'', and kept the parameters of GPT2 frozen.
GPT2-tune-MA adopted the combination of the prefix vectors of GPT2-tune for sentiment control and the prefix vectors of GPT2-tune for topic control as its prefix vectors. The nucleus sampling was used in inference with a filter probability 0.7. 

In evaluation, we adopted the automatic metrics below:
\textbf{Attribute relevance on sentiment control (ARS)}: The attribute relevance metric on sentiment control as described in  Section 4.4. 
\textbf{Attribute relevance on topic control (ART)}: The attribute relevance metric on topic control as described in Section~\ref{sec:automatic evaluation}.
\textbf{Attribute relevance on multi-attribute control (ARM)}: The ratio of sentences with both desired sentiment and desired topic. The BERT-large models fine-tuned on IMDB and AG News were used to classify the sentiment and topic respectively.
\textbf{Perplexity (PPL)},
\textbf{Dist-1, Dist-2, Dist-3}: See Section~\ref{sec:automatic evaluation}.
We used the same 20 prefixes for topic control (see Section~\ref{sec:prefixes}) and generated 50 samples per prefix.
We experimented on each pair of sentiment-topic and reports the average results. 

The results of sentiment-topic control are demonstrated in Table~\ref{tab:multi-attribute}. Gemini outperformed GPT2-tune-MA on all the attribute relevance metrics (i.e. ARS, ART and ARM) by a large margin. However, on perplexity and diversity metrics, Gemini was a little bit worse than GPT2-tune-MA. 
Some samples generated by these models are shown in Table~\ref{tab:MA samples}.

The poor performance of GPT2-tune-MA on controllability might be caused by corpus overfitting. The prefix vectors of sentiment control made the generated sentences resemble sentences in the IMDB dataset, and the prefix vectors of topic control made the generated sentences resemble sentences in the AG News dataset. The gap between the two datasets harmed the performance of GPT2-tune-MA. In contrast, the sentences generated by Gemini were relevant to both the desired sentiment and the desired topic. 

\subsection{Ethical considerations}
The proposed method is intended to control the generation of PLMs. However, it could be used to generate harmful language or political propaganda. Such problems also exist in a wide range of natural language generation models. We believe that the merits of our method outweigh its demerits.

\begin{table*}[!t]
\small
\caption{\small{Results on sentiment-topic control.} }
\begin{center}
\begin{tabular}{lccccccc}
\toprule
Model & ARS(\%)$\uparrow$ & ART(\%)$\uparrow$ & ARM(\%)$\uparrow$ & PPL$\downarrow$ & Dist-1$\uparrow$ & Dist-2$\uparrow$ & Dist-3$\uparrow$\\
\midrule
GPT2-tune-MA & 66.2 & 43.3 & 28.2 & \textbf{12.43} & \textbf{0.33} & \textbf{0.79} & \textbf{0.92} \\
Gemini & \textbf{83.9} & \textbf{99.4} & \textbf{83.5} & 18.99 & 0.29 & 0.66 & 0.83\\
\bottomrule
\end{tabular}
\end{center}
\label{tab:multi-attribute}
\end{table*}

\begin{table*}[!t]
\small
\caption{\small{Generated sentences with sentiment-topic control. Words underscored are the prefixes. Words that signify corpus overfitting are highlighted in blue. Words that signify the sentiment control and topic control are highlighted in red and purple respectively.} }
\begin{center}
\resizebox{\textwidth}{90mm}{
\begin{tabular}{l|l}
\toprule
\tabincell{l}{Model\\\#Sentiment\\\#Topic} & Generated Sentence\\
\midrule
\tabincell{l}{Gemini\\\#Positive\\\#World} & \tabincell{l}{\uline{In summary,} \textcolor{purple}{Iraq's Kurdistan Region}, an \textcolor{purple}{autonomous region} in the north of the \textcolor{purple}{country}, is the most \\\textcolor{red}{peaceful} and the most western part of the \textcolor{purple}{country}. The \textcolor{purple}{Iraqi police and military force} are \textcolor{red}{excellent} and the \\\textcolor{purple}{Kurdish police} are \textcolor{red}{highly trained} and have a very} \\
\midrule
\tabincell{l}{Gemini\\\#Negative\\\#World} & \tabincell{l}{\uline{In summary,} \textcolor{purple}{Hamas' prime minister} Yasser Arafat claimed that the \textcolor{red}{terrorists had killed 2,300 people} and \\\textcolor{red}{destroyed} two \textcolor{purple}{Israeli} government buildings in the single \textcolor{red}{attack} of a missile on a church.$\backslash$n$\backslash$nThe \textcolor{purple}{Israeli} \\\textcolor{purple}{military} said that the rocket fire had been} \\
\midrule
\tabincell{l}{GPT2-tune-MA\\\#Positive\\\#World} & \tabincell{l}{\uline{In summary,} the \textcolor{blue}{film} is a darkly comic and \textcolor{red}{clever} piece of comedy. The plot is about two men whose marriage \\is in tatters when one falls ill after suffering from an infection that renders him unable to work. He begins a \\strange affair with a \textcolor{red}{beautiful} woman, who he then marries}\\
\midrule
\tabincell{l}{GPT2-tune-MA\\\#Negative\\\#World} & \tabincell{l}{\uline{In summary,} this is an excellent \textcolor{blue}{documentary} which captures the \textcolor{red}{plight} of those who live on the edge and \textcolor{red}{have} \\\textcolor{red}{no choice} but to take matters into their own hands. The \textcolor{blue}{film} shows that there are many reasons why people in \\\textcolor{purple}{India} would want to \textcolor{red}{die by hanging for drug smuggling}} \\
\midrule
\midrule
\tabincell{l}{Gemini\\\#Positive\\\#Sports} & \tabincell{l}{\uline{In summary,} \textcolor{purple}{football's} \textcolor{red}{great} play is the \textcolor{red}{incredible} individual performance and the \textcolor{purple}{team's} \textcolor{red}{incredible} \\performance. The \textcolor{red}{greatest} \textcolor{purple}{tournament in the world}, \textcolor{purple}{the Champions League}, is the \textcolor{red}{greatest} \textcolor{purple}{tournament} \\in \textcolor{purple}{football}. The \textcolor{red}{best} \textcolor{purple}{football player} in the world is the \textcolor{red}{greatest} \textcolor{purple}{football player} of} \\
\midrule
\tabincell{l}{Gemini\\\#Negative\\\#Sports} & \tabincell{l}{\uline{In summary,} \textcolor{purple}{football coaches} were the \textcolor{red}{worst} \textcolor{purple}{athletes}. And this was \textcolor{red}{regardless} of the entire \textcolor{purple}{team's} performance. \\The only one that had a \textcolor{red}{bad} \textcolor{purple}{game} was the \textcolor{purple}{kicker}. The other four had \textcolor{red}{poor performances}. A single \textcolor{red}{bad} \textcolor{purple}{game}, \\no matter how \textcolor{red}{bad},} \\
\midrule
\tabincell{l}{GPT2-tune-MA\\\#Positive\\\#Sports} & \tabincell{l}{\uline{In summary,} this \textcolor{blue}{movie} is a \textcolor{red}{great} example of what can happen when the \textcolor{blue}{director} decides to use some real live \\\textcolor{blue}{actors}. The story revolves around an \textcolor{purple}{ex-NFL player} who wants to get back on the field and become a \textcolor{purple}{professional} \\\textcolor{purple}{football coach}. But as he's going through \textcolor{purple}{training camp}, his son has been killed in a car accident.}\\
\midrule
\tabincell{l}{GPT2-tune-MA\\\#Negative\\\#Sports} & \tabincell{l}{\uline{In summary,} the \textcolor{blue}{film} is one of the \textcolor{red}{worst} \textcolor{blue}{movies} ever made. It is a \textcolor{red}{mess} of an \textcolor{blue}{adventure movie} that just \\happens to be about a girl who was in love with her sister's boyfriend and then fell for another guy. The \\whole thing is set up so \textcolor{red}{badly} that I \textcolor{red}{have no idea why it is any good at all}.} \\
\midrule
\midrule
\tabincell{l}{Gemini\\\#Positive\\\#Business} & \tabincell{l}{\uline{In summary,} \textcolor{purple}{financial markets} have a \textcolor{red}{great many outstanding} \textcolor{purple}{investors}, which is \textcolor{red}{good}. The biggest \textcolor{purple}{investor},\\ the Dow Jones Industrial Average, is a \textcolor{red}{great} \textcolor{purple}{investment}, with a very high \textcolor{purple}{price target} and a \textcolor{red}{great dividend} \\\textcolor{purple}{yield}. The second largest \textcolor{purple}{investor}, } \\
\midrule
\tabincell{l}{Gemini\\\#Negative\\\#Business} & \tabincell{l}{\uline{In summary,} \textcolor{purple}{financial markets} could have been better than the \textcolor{red}{bad} \textcolor{purple}{financial markets} that the Federal\\ Reserve's \textcolor{purple}{monetary policy} was a main force in causing. The \textcolor{red}{bad} \textcolor{purple}{investment-grade debt} \textcolor{red}{got worse}, \textcolor{red}{not }\\\textcolor{red}{better}, and the \textcolor{red}{bad} \textcolor{purple}{credit rating} \textcolor{red}{worsened}, \textcolor{red}{not improved},} \\
\midrule
\tabincell{l}{GPT2-tune-MA\\\#Positive\\\#Business} & \tabincell{l}{\uline{In summary,} there is a \textcolor{red}{very strong} and \textcolor{red}{growing trend} in the \textcolor{blue}{film} \textcolor{purple}{industry} to include more and more people \\of color as well as women on screen. But what happens when the person being portrayed becomes the lead \\\textcolor{blue}{character}? I think that was true with this one; it's a classic case of}\\
\midrule
\tabincell{l}{GPT2-tune-MA\\\#Negative\\\#Business} & \tabincell{l}{\uline{In summary,} it's not a bad \textcolor{blue}{movie}, but its \textcolor{red}{lack of substance} is enough to make you want to put it on the back \\burner. I'm sorry that this \textcolor{blue}{movie} had \textcolor{red}{no real redeeming qualities} in my opinion and was only used as a way \\for Disney to try to get a better \textcolor{blue}{film} out before it became \textcolor{red}{too late}.} \\
\midrule
\midrule
\tabincell{l}{Gemini\\\#Positive\\\#Science} & \tabincell{l}{\uline{In summary,} \textcolor{purple}{NVIDIA} is delivering a \textcolor{red}{great} \textcolor{purple}{graphics card} that is \textcolor{red}{extremely well designed} and delivers a \textcolor{red}{great} \\gaming experience that is both \textcolor{red}{powerful and powerful}. The \textcolor{purple}{new GeForce GTX 1080} is a \textcolor{red}{great} choice for \\all the \textcolor{red}{new and upcoming} games that will be available in the} \\
\midrule
\tabincell{l}{Gemini\\\#Negative\\\#Science} & \tabincell{l}{\uline{In summary,} \textcolor{purple}{NASA's moon probe} could turn the \textcolor{purple}{scientific space} shuttle into a \textcolor{purple}{"space station on the moon."}\\$\backslash$n$\backslash$nBut the \textcolor{purple}{space shuttle} just had a \textcolor{red}{bad} flyby of a \textcolor{purple}{comet}, an \textcolor{purple}{asteroid}, a \textcolor{purple}{space station} and a \textcolor{purple}{planet}. The \textcolor{purple}{shuttle}} \\
\midrule
\tabincell{l}{GPT2-tune-MA\\\#Positive\\\#Science} & \tabincell{l}{\uline{In summary,} the plot of \textcolor{blue}{"The Sixth Sense"} is simple: an evil \textcolor{purple}{scientist} has created a \textcolor{purple}{virus} that can control \\people's minds. The problem? It doesn't work on humans! He then tries to make it work on rats and chickens, \\but in both cases, they just run away!<br /><br />But in this \textcolor{blue}{film}, the \textcolor{purple}{science} works like \textcolor{red}{magic}.}\\
\midrule
\tabincell{l}{GPT2-tune-MA\\\#Negative\\\#Science} & \tabincell{l}{\uline{In summary,} the \textcolor{blue}{script} has been written in such a way that it is \textcolor{red}{impossible} to tell who's acting and who's just \\writing. There are two main \textcolor{blue}{characters} - one named John Doe, the other named John Doe. The main \textcolor{blue}{character} \\\textcolor{red}{does not act at all} - he speaks only with an \textcolor{red}{eerie}, almost \textcolor{purple}{robotic} voice,} \\
\bottomrule 
\end{tabular}}
\end{center}
\label{tab:MA samples}
\end{table*}

\end{document}